\documentclass[conference]{IEEEtran}
\IEEEoverridecommandlockouts
\usepackage{cite}
\usepackage{amsmath,amssymb,amsfonts}
\usepackage{algorithmic}
\usepackage{graphicx}
\usepackage{textcomp}
\usepackage{xcolor}
\usepackage{multirow}
\usepackage{natbib}
\usepackage{amsthm}
\usepackage{comment}

\usepackage{booktabs}
\usepackage{amsmath}
\usepackage{mathtools}
\usepackage{amsfonts}
\usepackage{bm}
\usepackage{bbm}

\usepackage{algorithm}
\usepackage{algorithmic}

\def\eqref#1{equation~\ref{#1}}
\def\1{\bm{1}}

\DeclareMathAlphabet{\mathsfit}{\encodingdefault}{\sfdefault}{m}{sl}
\SetMathAlphabet{\mathsfit}{bold}{\encodingdefault}{\sfdefault}{bx}{n}

\DeclareMathOperator*{\argmin}{arg\,min}

\newcommand{\iid}{ \stackrel{\mathrm{i.i.d}}{\sim} }

\usepackage{multirow}

\newtheorem{theorem}{Theorem}
\newtheorem{lemma}[theorem]{Lemma}

\usepackage[textsize=tiny]{todonotes}

\renewcommand{\eqref}[1]{(\ref{#1})}

\def\BibTeX{{\rm B\kern-.05em{\sc i\kern-.025em b}\kern-.08em
    T\kern-.1667em\lower.7ex\hbox{E}\kern-.125emX}}
\begin{document}

\title{Learning from Double Positive and Unlabeled Data for Potential-Customer Identification}

\author{\IEEEauthorblockN{Masahiro Kato}
\IEEEauthorblockA{\textit{Mizuho-DL Financial Technology} \\
Tokyo, Japan\\
masahiro-kato@fintec.co.jp}
\and
\IEEEauthorblockN{Yuki Ikeda}
\IEEEauthorblockA{\textit{Mizuho-DL Financial Technology} \\
Tokyo, Japan\\
yuki-ikeda@fintec.co.jp}
\and
\IEEEauthorblockN{Kentaro Baba}
\IEEEauthorblockA{\textit{Mizuho-DL Financial Technology} \\
Tokyo, Japan\\
kentaro-baba@fintec.co.jp}
\and
\IEEEauthorblockN{Takashi Imai}
\IEEEauthorblockA{\textit{Mizuho-DL Financial Technology} \\
Tokyo, Japan\\
takashi-imai@fintec.co.jp}
\and
\IEEEauthorblockN{Ryo Inokuchi}
\IEEEauthorblockA{\textit{Mizuho-DL Financial Technology} \\
Tokyo, Japan\\
ryo-inokuchi@fintec.co.jp}
}
\maketitle

\begin{abstract}
In this study, we propose a method for identifying \emph{potential customers} in targeted marketing by applying \emph{learning from positive and unlabeled data} (PU learning). We consider a scenario in which a company sells a product and can observe only the customers who purchased it. Decision-makers seek to market products effectively based on whether people have loyalty to the company. Individuals with loyalty are those who are likely to remain interested in the company even without additional advertising. Consequently, those loyal customers would likely purchase from the company if they are interested in the product. In contrast, people with lower loyalty may overlook the product or buy similar products from other companies unless they receive marketing attention. Therefore, by focusing marketing efforts on individuals who are interested in the product but do not have strong loyalty, we can achieve more efficient marketing. To achieve this goal, we consider how to learn, from limited data, a classifier that identifies potential customers who (i) have interest in the product and (ii) do not have loyalty to the company. Although our algorithm comprises a single-stage optimization, its objective function implicitly contains two losses derived from standard PU learning settings. For this reason, we refer to our approach as \emph{double PU learning}. We verify the validity of the proposed algorithm through numerical experiments, confirming that it functions appropriately for the problem at hand.
\end{abstract}

\section{Introduction}
Data-driven, effective target marketing has attracted significant attention in business. To promote product sales, we focus on the set of \emph{potential customers} defined as:
\begin{itemize}
\item[(i)] people who might be interested in the product, but
\item[(ii)] do not have strong loyalty to the company.
\end{itemize}
This idea is based on the premise that loyal customers, who already feel an attachment to the company, are likely to purchase new products without additional advertising. On the other hand, those lacking strong company loyalty may fail to notice the product or choose similar products from competitors, even if they are interested. By identifying and targeting such potential customers, we aim to conduct marketing more efficiently.

In this work, we propose a framework for classifying individuals in a broader population to find those who meet the above definition of potential customers. Let each person be characterized by a feature vector $X$, and suppose they have two binary labels: 
\[
Y \in \{-1,+1\}\quad\text{and}\quad Z \in \{-1,+1\}.
\]
Here, $Y = +1$ indicates the person \emph{is interested in the product}, while $Z = +1$ indicates the person \emph{has loyalty to the company}. Conversely, $Y = -1$ and $Z = -1$ each signify the opposite of those attributes. We define a new binary label $W \in \{-1, +1\}$ such that
\begin{align*}
W = 
\begin{cases}
+1 & \text{if } Y = +1 \text{ and } Z = -1,\\
-1 & \text{otherwise}.
\end{cases}
\end{align*}
Thus, $W = +1$ identifies people who are \emph{potential customers} (interested in the product but lacking loyalty to the company), whereas $W = -1$ indicates they do not meet those conditions.

\begin{figure}[h]
\centering
\includegraphics[scale=0.3]{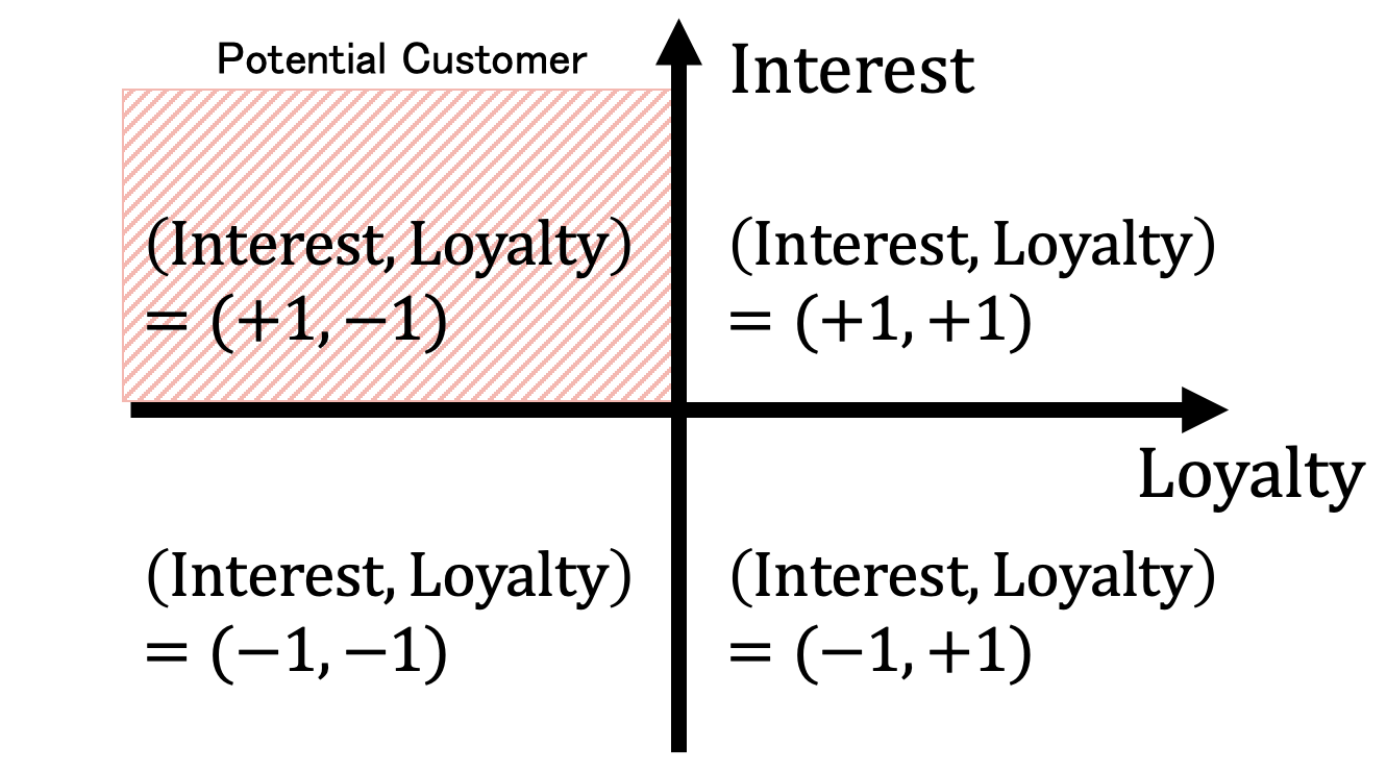}
\caption{Diagram illustrating how the interest label $Y$, loyalty label $Z$, and potential-customer label $W$ relate.}
\label{fig:fig4}
\end{figure}

Our goal is to learn a classifier that predicts $W$ using only \emph{positive data} and \emph{unlabeled data}. More concretely, we assume availability of three datasets:
\begin{itemize}
\item \textbf{Positive interest data} ($X \sim p(x \mid y =+1$)): 
  A set of people who have shown interest in the product, for instance by purchasing it or a similar item.
\item \textbf{Unlabeled data} ($X \sim p(x)$): 
  A dataset of individuals for whom the interest and loyalty labels are unknown. Only $X$ is observed.
\item \textbf{Positive loyalty data} ($X \sim p(x \mid y=+1, z=+1$)): 
  A set of people who have shown both interest in the product and loyalty to the company, for instance by repeatedly purchasing the company's products or registering for its services.
\end{itemize}
These datasets need not be strictly separate; some overlap among them is permissible. Figure~\ref{fig:fig4} illustrates how the interest label $Y$, the loyalty label $Z$, and the newly defined label $W$ for potential customers correspond.

In many real-world scenarios, it is difficult or expensive to obtain reliable negative labels: a lack of purchase does not necessarily prove \emph{no interest} in the product, and the absence of a subscription or account does not directly show \emph{no loyalty} to the company. Consequently, we lack explicit negative labels. However, standard supervised learning algorithms typically require both positive and negative examples. \emph{Learning from positive and unlabeled data (PU learning)} is a form of weak supervision that addresses this issue and has been employed in domains such as information retrieval, anomaly detection, and beyond \citep{elkan2008learning,ward2009presence,pmlr-v5-scott09a,blanchard2010semi,li2009positive,nguyen2011positive}.

Existing PU learning algorithms enable binary classification when only positive and unlabeled data are available. However, our setting involves two sorts of positive labels---one indicating product interest ($Y=+1$) and another indicating both product interest and company loyalty ($Y=+1, Z=+1$)---and we aim to distinguish $W=+1$ (potential customers) from $W=-1$. Hence, a direct application of standard PU learning methods is not sufficient.

We resolve this by extending existing PU learning techniques to incorporate the positive loyalty data. Concretely, we first define a loss function for classifying $Y=+1$ versus $Y=-1$. Since $W=+1$ corresponds to $Y=+1$ and $Z=-1$, we then adjust this loss function accordingly to construct a single objective function for predicting $W=+1$ versus $W=-1$. Despite being optimized in one stage, our approach effectively contains two PU learning losses: one for identifying those with $Y=+1$ and another for adjusting loyalty status ($Z=+1$ versus $Z=-1$). Hence, we refer to our approach as \emph{double PU learning}. Details of the proposed method are presented in Section~\ref{sec:dpu}.

In Section~\ref{sec:simulate}, we report simulation results to verify the validity of our algorithm. 
%追加
In Section~\ref{sec:marketing example}, we conduct an experiment using real-world marketing data. These experiments show that the proposed double PU learning classifier works as intended for the outlined task.

\section{Problem Setting}
\label{sec:prob}
Let $X\in\mathcal{X}\subset \mathbb{R}^d$ be a feature random variable that characterizes an individual. 
Each individual with $X$ potentially has two labels $Y\in\{-1, +1\}$ and $Z\in\{-1, +1\}$. In our setting, we can only observe labeled individuals with $Y = +1$ or $Z = +1$ or unlabeled individuals. Given the positive and unlabeled data, our interest lies in classifying $\{Y = +1\} \land \{Z = -1\}$ or not given $X$. We introduce more detailed setting below. 

\subsection{Potential Outcomes}
To deal with missing random variables, we introduce potential outcomes, which are used in the literature of missing outcomes \cite{Vaart1998} and causal inference \cite{Neyman1923,rubin1974estimating}.

Let $Y, Z \in \{-1, +1\}$ be binary labels, and $X\in\mathcal{X}\subset \mathbb{R}^d$ be a $d$-dimensional feature vector with a feature space $\mathcal{X}$. We suppose that the triple $(Y, Z, X)$ is generated as follows:
\begin{align*}
    (Y, Z, X) \sim p(y, z, x).
\end{align*}

In this study, we consider a situation where there are two different labels $Y$ and $Z$ in each object whose feature is represented by $X$. Specifically, in our application, we consider the following situation. 

\paragraph{Example:} Each object is a customer whose feature is summarized as $X$, such as its age. For the customer, $Y$ corresponds to an actual outcome of advertisement; that is, the customer bought a product or not after an advertisement is treated. Here, $Z$ corresponds to a loyalty of the customer, which represents that the customer likes the brand or not. If the labels are $Y = +1$ and $Z = +1$, the customer likes the brand and bought its product as a result of advertisement. If the labels are $Y = +1$ and $Z = -1$, the customer is not interested in the brand but bought its product.

\subsection{Observed Data}
\label{sec:observeddata}
This study assumes that we can observe the following three datasets, which are generated from $p\big(x\mid  y = +1\big)$, $p\big(x\big)$, and $p\big(x\mid (y, z)=(+1, +1)\big)$, respectively:
\begin{align*}
    \mathcal{D}_{y=+1} &\coloneqq \big\{X_j\big\}^J_{j=1} \iid p\big(x\mid  y = + 1\big),\\
    \mathcal{D}_{U} &\coloneqq \big\{X_k\big\}^K_{k=1} \iid p\big(x\big),\\
    \mathcal{D}_{(y, z) = (+1, +1)} &\coloneqq \big\{X_l\big\}^L_{l=1} \iid p\big(x\mid (y, z)=(+1, +1)\big).
\end{align*}
Here, we defined $\mathcal{D}_{(y, z) = (+1, +1)}$ as a dataset independent of $\mathcal{D}_{y=+1}$.
However, $\mathcal{D}_{(y, z) = (+1, +1)}$ can be observed as a subset of $\mathcal{D}_{y=+1}$; that is, $\mathcal{D}_{(y, z) = (+1, +1)}\subset \mathcal{D}_{y=+1}$. 

This setting is a variant of case-control PU learning \cite{Gang2016}, which has been adopted by \cite{duPlessis2015}, \cite{Kiryo2017}, and \cite{kato2018pubp}. Another framework, known as censoring PU learning, has been used in earlier studies such as \cite{elkan2008learning} and \cite{Bekker2018}. Although case-control PU learning accommodates a more general setting than censoring PU learning, it requires additional information to train a classifier.

\subsection{Classification Risk and Optimal Classifiers}
\begin{figure}[h]\centering
\includegraphics[scale=0.3]{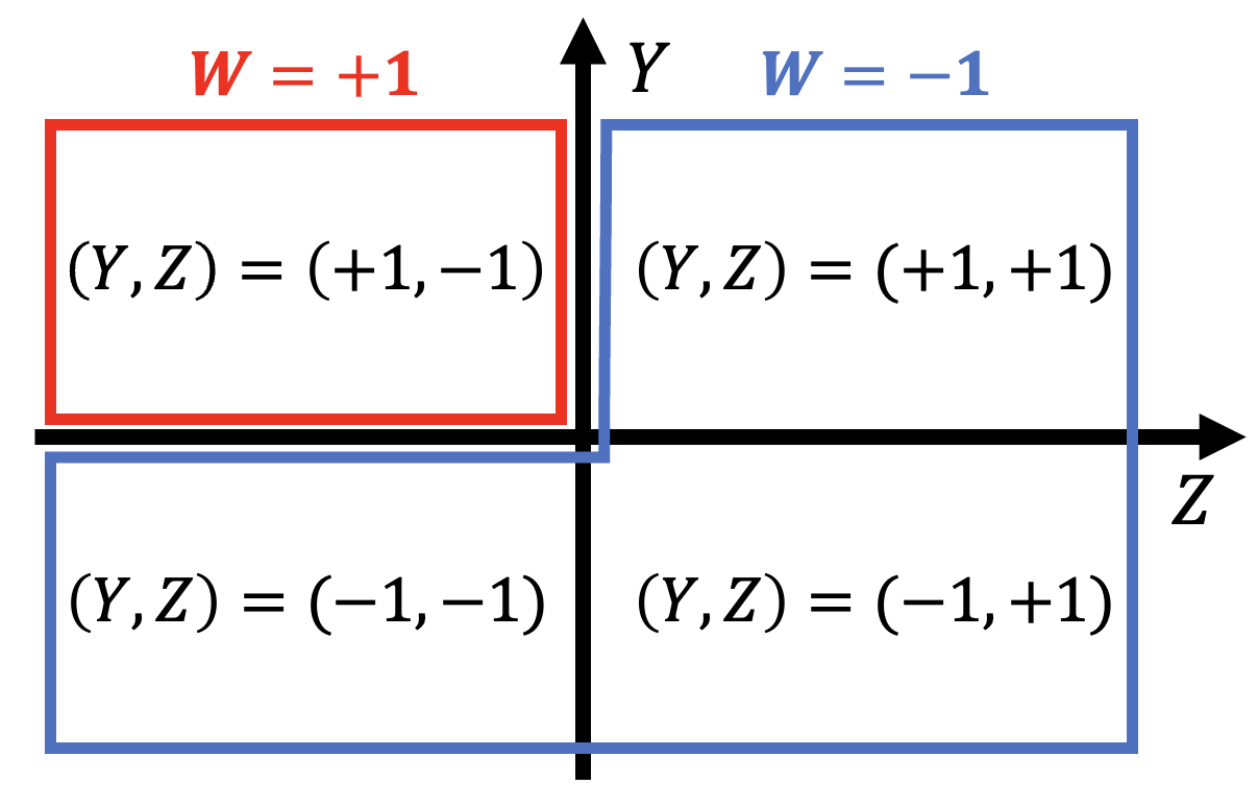}
\caption{Quadrants divided by $Y$ and $Z$. The upper left quadrant represents $W=+1$, and the other quadrants are $W=-1$.}
\label{fig:fig1}
\end{figure}

Our interest lies in classifying $(Y, Z) = (+1, -1)$ and not $(Y, Z) = (+1, -1)$ given $X$ (Figure~\ref{fig:fig1}); that is, we define positive label $W = + 1$ as $(Y,Z)=(+1, -1)$ and negative label $W = - 1$ as $\big\{(Y,Z)=(+1, +1)\big\}\cup \big\{(Y,Z)=(-1, +1)\big\}\cup \big\{(Y,Z)=(-1, -1)\big\}$:
\begin{align*}
    &\mathrm{positive:}\ W = +1\ \mathrm{if}\ (Y,Z)=(+1, -1),\\
    &\mathrm{negative:}\ W = -1\ \mathrm{otherwise}.
\end{align*}
Here, we note that because we cannot observe which observations are labeled $Y = -1$ or $Z = -1$. Figure~\ref{fig:fig1} shows the quadrants divided by these labels. The upper left quadrant represents $W=+1$, and the other quadrants are $W=-1$.

\paragraph{Classification risk.}
Let $g:\mathcal{X} \to \mathbb{R}$ be a classifier and $\mathcal{G}$ be some set of $g$. Let $\mathbb{E}_{W = +1}$ and $\mathbb{E}_{W = -1}$ be expectations over $p\big(x\mid w = +1\big) = p\big(x\mid (y, z) = (+1, -1)\big)$ (density of $X$ conditioned on a positive label) and $p\big(x\mid w = -1\big) = p\big(x\mid (y, z) = (+1, +1) \lor (-1, +1) \lor (-1, -1)\big)$ (density of $X$ conditioned on a negative label). Then, we define an optimal classifier as 
\[g^* = \argmin_{g\in\mathcal{G}}R_{0\mathchar`-1}(g),\]
where $R_{0\mathchar`-1}(g)$ is the expected misclassification rate when the classifier $g(X)$ is applied to marginal (unlabeled) feature distribution $p(x)$, defined as
\begin{align*}
    R_{0\mathchar`-1}(g) =& \alpha\mathbb{E}_{W=+1}\big[\ell_{0\mathchar`-1}(g(X))\big] \\
    &+ (1-\alpha)\mathbb{E}_{W=-1}\big[\ell_{0\mathchar`-1}(-g(X))\big],
\end{align*}
where $\ell_{0\mathchar`-1}(z) = \frac{1}{2}\mathrm{sgn(-z)} + \frac{1}{2}$ is the zero-one loss and $\alpha = p(w = +1) = p\big(y = +1, z = -1\big)$ is called the class-prior. We also define $\beta = p(y = +1)$ and $\gamma = p(y = +1, z= +1)$ and assume that $\beta$ and $\gamma$ are known.

\section{Double PU Learning}
\label{sec:dpu}
This section provides our algorithm. First, we discuss an unbiased risk for $R_{0\mathchar`-1}(g)$ in Section~\ref{sec:unbiased_risk}. Next, we define empirical risk minimization in Section~\ref{sec:empirical_risk}. 

\subsection{Identification}
If we have data sampled from $p(x\mid w = +1)$ and $p(x\mid w = -1)$, we can directly construct $R_{0\mathchar`-1}(g) = \alpha\mathbb{E}_{W=+1}\big[\ell_{0\mathchar`-1}(g(X))\big] + (1-\alpha)\mathbb{E}_{W=-1}\big[\ell_{0\mathchar`-1}(-g(X))\big]$. However, since we do not assume the access to the data, we consider another approach.

Here, we discuss how we identify $R_{0\mathchar`-1}(g)$ from observations.
For a function $h:\mathcal{X}\to \mathbb{R}$, we define the expected value of $h(X)$ over $p(x\mid y = +1)$, $p(x)$, and $p(x\mid y = +1, z = +1)$ as $\mathbb{E}_{y=+1}[h(X)]$, $\mathbb{E}_{U}[h(X)]$, and $\mathbb{E}_{y=+1, z = +1}[h(X)]$, respectively. Then, we write $R_{0\mathchar`-1}(g)$ using $\mathbb{E}_{y=+1}[h(X)]$, $\mathbb{E}_{U}[h(X)]$, and $\mathbb{E}_{y=+1, z = +1}[h(X)]$ as follows:
\begin{lemma}
\label{lemma:loss}
We have
\begin{align*}
    R_{0\mathchar`-1}(g) &= \beta \mathbb{E}_{Y=+1}\big[\ell_{0\mathchar`-1}(g(X))\big]\\
    &\ \ \ - \gamma \mathbb{E}_{(Y, Z) = (+1, +1)}\big[\ell_{0\mathchar`-1}(g(X))\big] + \mathbb{E}_{U}\big[\ell_{0\mathchar`-1}(-g(X))\big]\\
    &\ \ \ - \beta\mathbb{E}_{Y=+1}\big[\ell_{0\mathchar`-1}(-g(X))\big]\\
    &\ \ \ + \gamma\mathbb{E}_{(Y, Z) = (+1, +1)}\big[\ell_{0\mathchar`-1}(-g(X))\big],
\end{align*}
where $\beta \coloneqq p(y = +1)$ and $\gamma \coloneqq p(y = +1, z = +1)$. 
\end{lemma}
\begin{proof}
We have
\begin{align*}
    R_{0\mathchar`-1}(g) &= \alpha\mathbb{E}_{W=+1}\big[\ell_{0\mathchar`-1}(g(X))\big]\\
    &\ \ \ + (1-\alpha)\mathbb{E}_{W=-1}\big[\ell_{0\mathchar`-1}(-g(X))\big]\\
    &= \alpha\mathbb{E}_{W=+1}\big[\ell_{0\mathchar`-1}(g(X))\big]\\
    &\ \ \ + \mathbb{E}_{U}\big[\ell_{0\mathchar`-1}(-g(X))\big] - \alpha\mathbb{E}_{W=+1}\big[\ell_{0\mathchar`-1}(-g(X))\big]\\
    &= \beta \mathbb{E}_{Y=+1}\big[\ell_{0\mathchar`-1}(g(X))\big]\\
    &\ \ \ - \gamma \mathbb{E}_{(Y, Z) = (+1, +1)}\big[\ell_{0\mathchar`-1}(g(X))\big] + \mathbb{E}_{U}\big[\ell_{0\mathchar`-1}(-g(X))\big]\\
    &\ \ \ - \beta\mathbb{E}_{Y=+1}\big[\ell_{0\mathchar`-1}(-g(X))\big]\\
    &\ \ \ + \gamma\mathbb{E}_{(Y, Z) = (+1, +1)}\big[\ell_{0\mathchar`-1}(-g(X))\big].
\end{align*}
Here, we used 
\begin{align*}
    &\alpha p(x\mid w= +1) = p(x, w = +1) = p(x, y = +1, z = -1)\\
    &\ \ \ = p(x, y = +1) - p(x, y = +1, z = +1)\\
    &\ \ \ = \beta p(x\mid y = +1) - \gamma p(x\mid y = +1, z = +1).
\end{align*}
\end{proof}

\subsection{Unbiased Risk} 
\label{sec:unbiased_risk}
Thus, we obtain an equivalent formulation of $R_{0\mathchar`-1}(g)$ using $\mathbb{E}_{Y=+1}[h(X)]$, $\mathbb{E}_{U}[h(X)]$, and $\mathbb{E}_{(Y, Z) = (+1, +1)}[h(X)]$. This section presents an unbiased risk for $R_{0\mathchar`-1}(g)$ with replacing its zero-one loss with a surrogate loss. 

\paragraph{Sample approximation} 
For $h:\mathcal{X}\to \mathbb{R}$, we define sample approximations of $\mathbb{E}_{Y=+1}[h(X)]$, $\mathbb{E}_{U}[h(X)]$, and $\mathbb{E}_{(Y, Z) = (+1, +1)}[h(X)]$ as
\begin{align*}
    &\widehat{\mathbb{E}}_{Y=+1}[h(X)] = \frac{1}{J}\sum^J_{j=1}h(X_j),\\
    &\widehat{\mathbb{E}}_{U}[g(X)] = \frac{1}{K}\sum^K_{k=1}h(X_k),\\
    &\widehat{\mathbb{E}}_{(Y, Z) = (+1, +1)}[h(X)] = \frac{1}{L}\sum^L_{l=1}h(X_l).
\end{align*}

\paragraph{Unbiased empirical risk.}
Then, an unbiased risk is given as
\begin{align*}
    \widehat{R}_{0\mathchar`-1}(g) &\coloneqq \beta\widehat{\mathbb{E}}_{Y = +1}\Big[\ell_{0\mathchar`-1}\big(g(X)\big)\Big]\\
    &\ \ \ - \gamma \widehat{\mathbb{E}}_{(Y,Z)=(+1,+1)}\Big[\ell_{0\mathchar`-1}\big(g(X)\big)\Big]\\
    &\ \ \ + \widehat{\mathbb{E}}_{U}\Big[\ell_{0\mathchar`-1}\big(-g(X)\big)\Big]- \beta\widehat{\mathbb{E}}_{Y = +1}\Big[\ell_{0\mathchar`-1}\big(-g(X)\big)\Big]\\
    &\ \ \ + \gamma \widehat{\mathbb{E}}_{(Y,Z)=(+1,+1)}\Big[\ell_{0\mathchar`-1}\big(-g(X)\big)\Big].
\end{align*}
Here, it holds that $\mathbb{E}\big[\widehat{R}_{0\mathchar`-1}(g)\big] = R_{0\mathchar`-1}(g)$.

\subsection{Empirical Risk Minimization}
\label{sec:empirical_risk}
We replace the zero-one loss $\ell_{0\mathchar`-1}(z)$ with a surrogate loss $\ell:\mathcal{R} \to \mathbb{R}$.

\paragraph{Surrogate loss.} Since it is not easy to optimize the zero-one loss $\ell_{0\mathchar`-1}$ directly, we replace it with a surrogate loss, including the log-loss and hinge loss. Let us denote a surrogate loss by $\ell:\mathcal{R} \to \mathbb{R}$. We raise candidates of a surrogate loss in Table~\ref{tab:surroate}. From the computational perspective, the use of surrogate losses with convexity is preferable. However, there are applications where non-convex surrogate losses perform well. Using a surrogate loss $\ell$, we define the population risk as 
\begin{align*}
    R(g) = \alpha\mathbb{E}_{W=+1}\big[\ell(g(X))\big] + (1-\alpha)\mathbb{E}_{W=-1}\big[\ell(-g(X))\big].
\end{align*}

\begin{table}[t]
\caption{Surrogate loss $\ell:\mathcal{Z}\to \mathbb{R}^+$.}
    \centering
    \begin{tabular}{|c|c|c|}
    \hline
        Loss name & $\ell(z)$ & $\mathcal{Z}$ \\
        \hline
        Squared loss & $(z - 1)^2$ & $\mathbb{R}$ \\
        Log loss & $-\log(z)$ & $(0, 1)$ \\
        Logistic loss & $\log\big(1 + \exp(-z)\big)$ & $\mathbb{R}$ \\
        Hinge loss & $\max\big(0, 1-z\big)$ & $\mathbb{R}$ \\
        \hline
    \end{tabular}
    \label{tab:surroate}
\end{table}

\paragraph{Empirical risk minimization.}

We train $g$ as
\begin{align}
\label{eq:objective}
    \widehat{g} = \argmin_{g\in\mathcal{G}} \widehat{R}(g),
\end{align}
where
\begin{align*}
    \widehat{R}(g) &\coloneqq \beta\widehat{\mathbb{E}}_{Y = +1}\Big[\ell\big(g(X)\big)\Big]\\
    &\ \ \ + \widehat{\mathbb{E}}_{U}\Big[\ell\big(-g(X)\big)\Big] - \beta\widehat{\mathbb{E}}_{Y = +1}\Big[\ell\big(-g(X)\big)\Big] \\
    &\ \ \ - \gamma \widehat{\mathbb{E}}_{(Y,Z)=(+1,+1)}\Big[\ell\big(g(X)\big)\Big]\\
    &\ \ \ + \gamma \widehat{\mathbb{E}}_{(Y,Z)=(+1,+1)}\Big[\ell\big(-g(X)\big)\Big].
\end{align*}

We refer to an algorithm that trains a binary classifier by \eqref{eq:objective} as double PU learning, since this objective function essentially consists of two PU learning risks as explained in the following section. 

\subsection{Intuition behind our Algorithm}
We can decompose $\widehat{R}(g) $ as 
\begin{align*}
    \widehat{R}(g) &= \widehat{R}^{\mathrm{PU}}(g) - \gamma \widehat{\mathbb{E}}_{(Y,Z)=(+1,+1)}\Big[\ell\big(g(X)\big)\Big]\\
    &\ \ \ + \gamma \widehat{\mathbb{E}}_{(Y,Z)=(+1,+1)}\Big[\ell\big(-g(X)\big)\Big],
\end{align*}
where $\widehat{R}^{\mathrm{PU}}(g)$ is the standard PU learning risk defined as
\begin{align*}
    \widehat{R}^{\mathrm{PU}}(g) &\coloneqq \beta\widehat{\mathbb{E}}_{Y = +1}\Big[\ell\big(g(X)\big)\Big]\\
    &  + \widehat{\mathbb{E}}_{U}\Big[\ell\big(-g(X)\big)\Big] - \beta\widehat{\mathbb{E}}_{Y = +1}\Big[\ell\big(-g(X)\big)\Big].
\end{align*}

\begin{figure}[h]\centering
\includegraphics[scale=0.3]{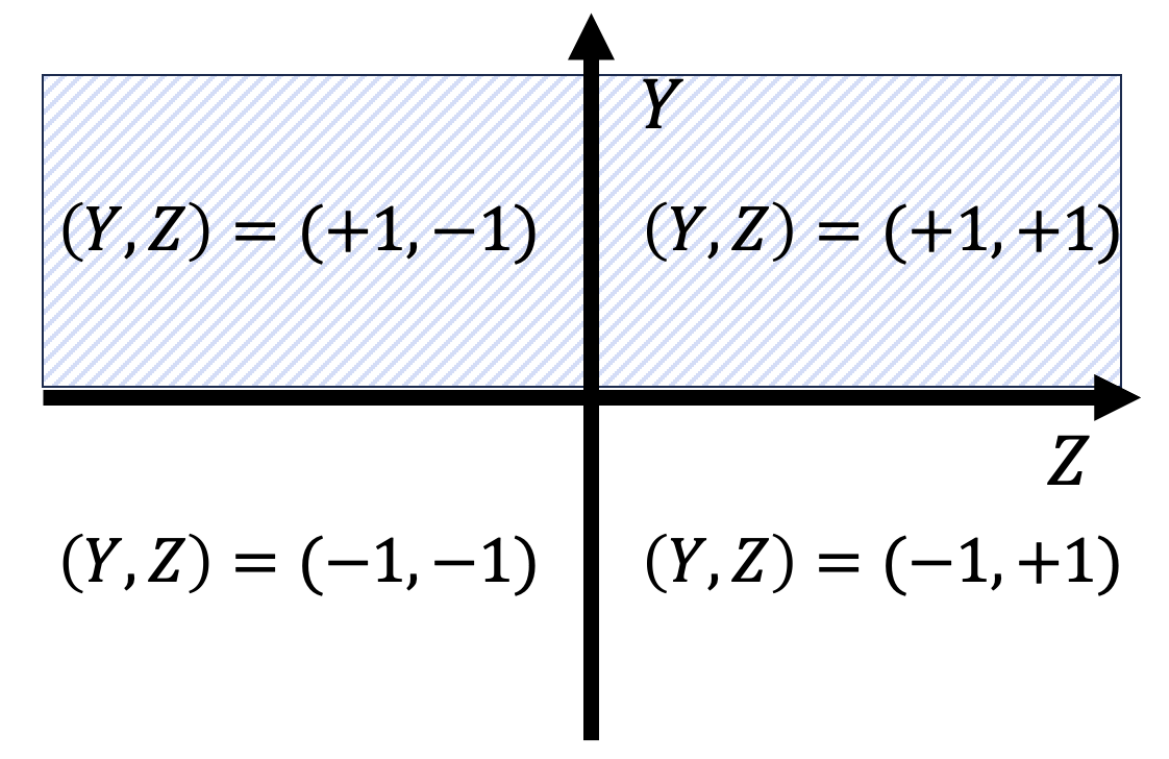}
\caption{The standard PU classification.}
\label{fig:fig2}
\end{figure}

\paragraph{Unbiased PU risk.}
The part $\widehat{R}^{\mathrm{PU}}(g)$ corresponds to an unbiased PU risk for classifying $Y = +1$ and $Y = -1$. By minimizing $\widehat{R}^{\mathrm{PU}}(g)$, we can train a classifier that classifies individuals into $Y = +1$ and $Y = -1$. Figure~\ref{fig:fig2} illustrates this classification, where the blue part corresponds to $Y = +1$ and the remaining part corresponds to $Y = -1$.

\begin{figure}[h]\centering
\includegraphics[scale=0.3]{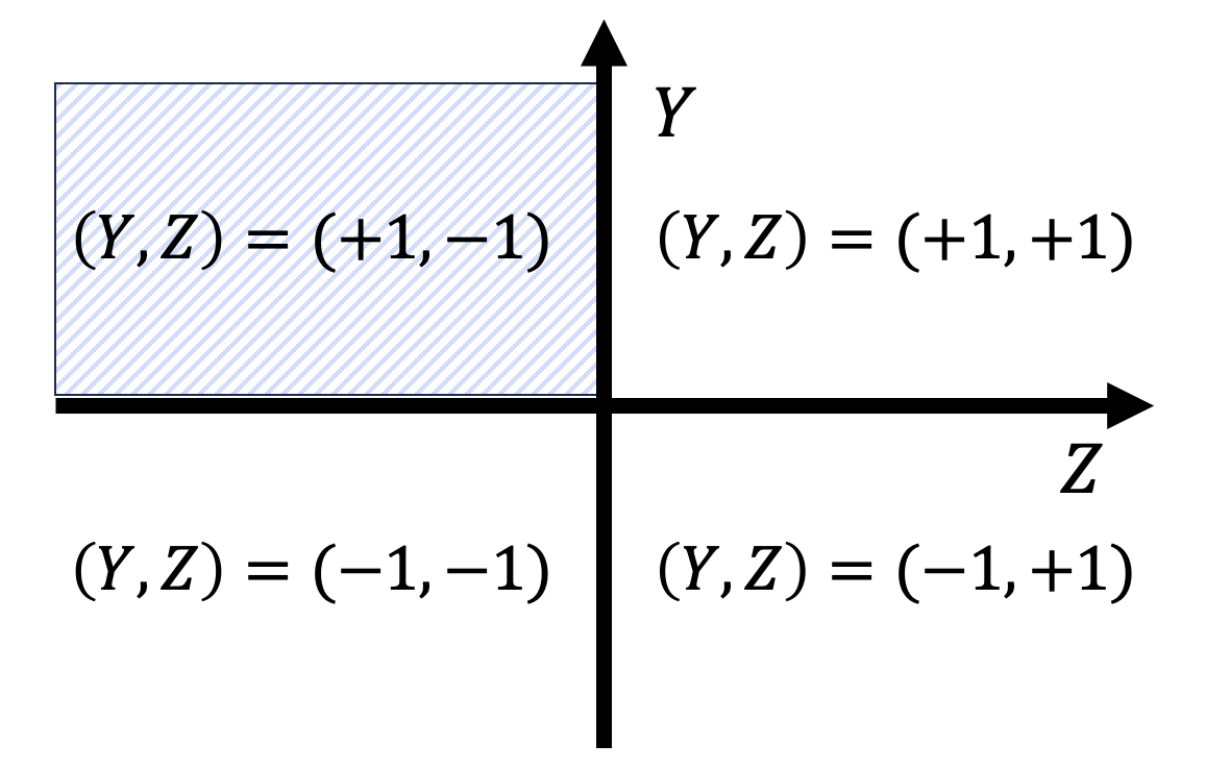}
\caption{The double PU classification.}
\label{fig:fig3}
\end{figure}

\paragraph{Unbiased double PU risk.}
The blue region in Figure~\ref{fig:fig2} includes $(Y, Z) = (+1, +1)$ in addition to $(Y, Z) = (+1, -1)$ ($W = + 1$). We aim to remove this region. This step can be interpreted as the second PU classification, since we minimize another empirical risk by regarding $Y = +1$ as unlabeled data and $(Y, Z) = (+1, +1)$ as labeled data. 
%This additional PU risk is given as
%\begin{align}
%    &\gamma \widehat{\mathbb{E}}_{(Y,Z)=(+1,+1)}\Big[\ell_{0\mathchar`-1}\big(-g(X)\big)\Big]\\
%    &- \gamma \widehat{\mathbb{E}}_{(Y,Z)=(+1,+1)}\Big[\ell_{0\mathchar`-1}\big(g(X)\big)\Big] + \beta\widehat{\mathbb{E}}_{Y = +1}\Big[\ell_{0\mathchar`-1}\big(g(X)\big)\Big].
%\end{align}
As a result, we obtain a classifier that classifies individuals into $W = +1$ and $W = -1$. In Figure~\ref{fig:fig3}, the blue part corresponds to $W = +1$ and the remaining part corresponds to $W = -1$.

Thus, our double PU learning algorithm trains a classifier for identifying potential customers by minimizing the unbiased empirical risk. This unbiasedness is important because it guarantees finite-sample stability of the learning process. In our setting, obtaining an unbiased risk estimator is not obvious, as the sampling scheme is non-standard in binary classification.

%\paragraph{On the use of neural networks.}
\subsection{Non-negative risk correction.}
When the hypothesis class $\mathcal{G}$ is highly complex—such as when it includes neural networks—empirical risk minimization in PU learning often breaks down \cite{Kiryo2017}. The problem stems from the negative term in the risk expression, which can diverge to negative infinity during training, a phenomenon referred to as train-loss hacking in \cite{Kato2021bregman}. To address this issue, \cite{Kiryo2017} introduced the non-negative risk correction, and the same correction can be applied to double PU learning
%スペースの関係で削除
%when neural networks are used
. This adjustment has been shown to enhance training stability.
%レビュアコメントに基づき追記
One way of the correction is
\begin{align*}
    & \widehat{R}(g) \coloneqq \beta\widehat{\mathbb{E}}_{Y = +1}\Big[\ell\big(g(X)\big)\Big]\\
    &\ \ \ + \max \Big\{0, \widehat{\mathbb{E}}_{U}\Big[\ell\big(-g(X)\big)\Big] - \beta\widehat{\mathbb{E}}_{Y = +1}\Big[\ell\big(-g(X)\big)\Big] \Big\}\\
    &\ \ \ - \gamma \widehat{\mathbb{E}}_{(Y,Z)=(+1,+1)}\Big[\ell\big(g(X)\big)\Big]\\
    &\ \ \ + \gamma \widehat{\mathbb{E}}_{(Y,Z)=(+1,+1)}\Big[\ell\big(-g(X)\big)\Big].
\end{align*}

%レビュアコメントに基づき追記
\subsection{Cost-sensitive learning.}
In many real applications, dataset is imbalanced in terms of class distribution, and cost-sensitive learning is one way to cope with this problem. Double-PU learning can be transformed to cost-sensitive setting. Slightly modifying the derivation of Lemma \ref{lemma:loss}, we suggest
\begin{align*}
    \widehat{R}(g) &\coloneqq c_{FN} \Big( \beta\widehat{\mathbb{E}}_{Y = +1}\Big[\ell\big(g(X)\big)\Big]\\
    &\ \ \ - \gamma \widehat{\mathbb{E}}_{(Y,Z)=(+1,+1)}\Big[\ell\big(g(X)\big)\Big] \Big)\\
    &\ \ \ + c_{FP} \Big(\widehat{\mathbb{E}}_{U}\Big[\ell\big(-g(X)\big)\Big] - \beta\widehat{\mathbb{E}}_{Y = +1}\Big[\ell\big(-g(X)\big)\Big] \\
    &\ \ \ + \gamma \widehat{\mathbb{E}}_{(Y,Z)=(+1,+1)}\Big[\ell\big(-g(X)\big)\Big] \Big),
\end{align*}
where $c_{FN}$ and $c_{FP}$ are the costs of false negative and false positive respectively. 
%We leave sensitivity analysis of the effect of imbalanced samples for future work.

\section{Simulation Studies}
\label{sec:simulate}
\begin{figure}[h]\centering
\includegraphics[scale=0.5]{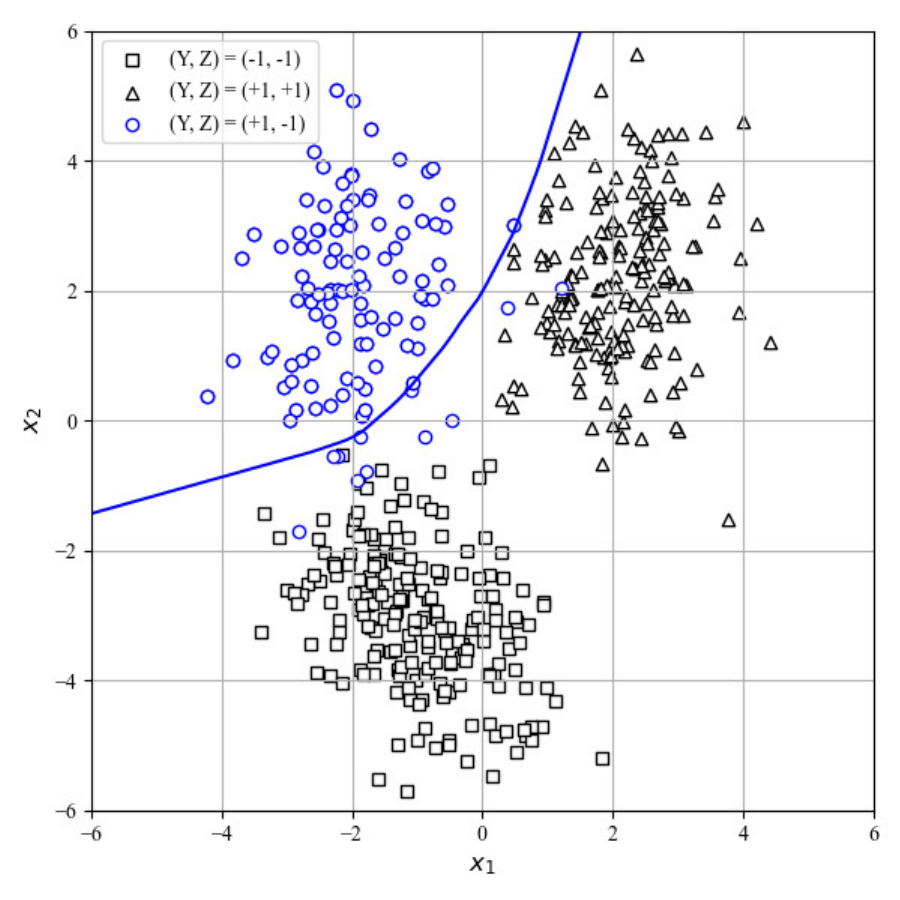}
\caption{Results of the numerical experiment.}
\label{fig:fig5}
\end{figure}

To verify the proposed algorithm we generated 500 positive samples with label $(Y,Z)=(+1,-1)$ and 1000 negative samples with labels $(Y,Z)=(+1,+1)$ and $(Y,Z)=(-1,-1)$, respectively, from bivariate Gaussian distributions (Figure~\ref{fig:fig5}).  
Using 70\% of the positive samples as labeled data, we randomized 30\% of both $(Y,Z)=(+1,-1)$ and $(Y,Z)=(+1,+1)$ to unlabeled for $Y$, and further randomized 50\% of $(Y,Z)=(+1,+1)$ to unlabeled for $Z$.  
Samples with $Y=-1$ were fully unlabeled for both $Y$ and $Z$.  
With $(x_1,x_2)$ as features we trained a logistic‑loss classifier.  
The learned decision boundary $p(w=+1\mid x)=0.5$ is shown in Figure~\ref{fig:fig5}, demonstrating that double PU learning successfully separates potential customers despite partial labeling.

%レビュアコメントに基づき追記
\section{Application to marketing}
\label{sec:marketing example}
We applied our method to bank marketing dataset \footnote{https://www.kaggle.com/datasets/janiobachmann/bank-marketing-dataset/data} originally from UCI machine learning repository.
There are 11,162 samples total with 15 features.
We want to predict customers who are likely to accept a marketing offer but not likely to default. In order to do so, we defined $Y=+1$ if a customer accepted a marketing offer and $Z=+1$ if the customer is in default. We want to find customers of $W=+1$ ($Y=+1$ but $Z=-1$). Note that this situation is different from the main motivation of this paper (finding a customer who have shown interest and low loyalty), but this setting also aligns with double-PU learning, because a bank often does not know a customer defaults unless the bank itself offers a loan to the customer.  
%Categorical features were transformed to numerical features using one-hot encoding.
%We would like to leave an experiment using real-world marketing data with higher dimension as a future work.
We first divided the original dataset into training and test dataset by 80\% and 20\%. Next, we further divided the training dataset into three datasets where the first one (10\% of total) used only samples with $Y=+1$ (corresponding to $\mathcal{D}_{y=+1}$), the second one (10\% of total)
used all samples (corresponding to $\mathcal{D}_{U}$) and the third one (80\% of total) used only samples with $Y=+1$ and $Z=+1$ (corresponding to $\mathcal{D}_{(y,z)=(+1,+1)}$).
We applied double-PU learning to these three datasets.
We set $\beta=0.4738$ and $\gamma=0.0046$ and employed a linear classifier for $g$.
Also, since a customer default can result in a significant cost, we assigned a cost to false positives that is $100$ times higher than the cost of false negatives for predicting $W$ using the cost-sensitive learning explained above.
As a result of the experiment, ROC-AUC of the learned classifier on the test dataset is 0.6013, implying that it can predict better than the random classifier.

\section{Related Work}
PU learning has also been investigated in statistics and economics \cite{Lancaster1996casecontrol}. In the machine-learning literature, \cite{elkan2008learning} proposed a practical algorithm for PU learning, and \cite{duPlessis2015} later introduced another practical approach, known as unbiased PU learning.

%レビュアコメントに基づき追記
We assumed the class priors $\beta$ and $\gamma$ are known, but this is often not the case in real applications. While the class prior can be estimated under certain conditions \cite{Kato2018alternateestimationclassifierclassprior}, doing so introduces extra assumptions. 
%We leave sensitivity analysis of the effect of class prior estimation on the double-PU learning performance for future work.

Two main data-generation mechanisms are considered in PU learning: censoring PU learning and case-control PU learning (see Section~\ref{sec:observeddata}). The setting analyzed by \cite{elkan2008learning} falls under censoring PU learning, whereas those in \cite{Lancaster1996casecontrol} and \cite{duPlessis2015} correspond to case-control PU learning. In case-control PU learning, positive instances are gathered independently of the unlabeled sample. In contrast, in censoring PU learning, a single sample is drawn first, after which only a subset of its positive instances is revealed. Due to this property, censoring PU learning is also called the one-sample scenario, while case-control PU learning is also called the two-sample scenario. 

Our problem setting is a variant of case-control PU learning, but it can also be formulated within the censoring PU learning framework. Although the case-control setting is slightly more general, it requires knowledge of the class prior to train the classifier. 
%上に移動
%While the class prior can be estimated under certain conditions \cite{Kato2018alternateestimationclassifierclassprior}, doing so introduces extra assumptions.
As a result, censoring PU learning algorithms are sometimes viewed as more practical. 
%スペースの関係で削除
%We omit a detailed discussion of that version here due to space constraints. 
Note that our algorithm is directly applicable to the censoring PU setting\footnote{As \cite{Gang2016} discusses, most case-control PU algorithm can be applied to censoring PU learning setting when the class prior is known}. 

%As elucidated by \cite{elkan2008learning}, the identification of a classifier hinges on an assumption regarding the labeling process of the positive data. The assumption traditionally made is that of \emph{selected completely at random} (SCAR), implying that the distribution of the labeled positive data is identical to that of the unlabeled positive data \cite{elkan2008learning, duPlessis2015}. 

%We illustrate the case of weighting samples following the framework introduced by \cite{elkan2008learning} in the appendix.

\section{Conclusion}
In this study, we proposed the double PU learning approach for targeted marketing. Our focus was on identifying potential customers who are interested in a product but do not have loyalty to the company. We proposed an algorithm for training a binary classifier in this task, despite limited data availability, referred to as the double PU learning. We confirmed the soundness of the proposed algorithm via simulated studies.

\bibliography{conference_041818.bbl}
\bibliographystyle{plain}

\end{document}